\documentclass[letterpaper]{article} 
\usepackage{aaai25}  
\usepackage{times}  
\usepackage{helvet}  
\usepackage{courier}  
\usepackage[hyphens]{url}  
\usepackage{graphicx} 
\usepackage{natbib}  
\usepackage{caption} 
\usepackage{algorithm}

\usepackage{color}
\usepackage{xcolor}
\usepackage{bbold}
\usepackage{amsmath}
\usepackage{algpseudocode}
\usepackage{multirow}

\usepackage{newfloat}
\usepackage{listings}
\DeclareCaptionStyle{ruled}{labelfont=normalfont,labelsep=colon,strut=off} 
\floatstyle{ruled}
\newfloat{listing}{tb}{lst}{}
\floatname{listing}{Listing}
\title{Cross-Patient Pseudo Bags Generation and Curriculum Contrastive Learning for Imbalanced Multiclassification of Whole Slide Image}
\author {
    Yonghuang Wu \textsuperscript{\rm 1},
    Xuan Xie \textsuperscript{\rm 1},
    Xinyuan Niu \textsuperscript{\rm 1},
    Chengqian Zhao \textsuperscript{\rm 1},
    Jinhua Yu \textsuperscript{\rm 1}\thanks{Corresponding author. Email: jhyu@fudan.edu.cn},
}
\affiliations {
    \textsuperscript{\rm 1}Fudan University\\
}

\definecolor{clcl}{RGB}{0,0,0}
\definecolor{clcl2}{RGB}{0,0,0}   

\usepackage{bibentry}

\begin{document}

\maketitle

\begin{abstract}
Pathology computing has dramatically improved pathologists' workflow and diagnostic decision-making processes. Although computer-aided diagnostic systems have shown considerable value in whole slide image (WSI) analysis, the problem of multi-classification under sample imbalance remains an intractable challenge. To address this, we propose learning fine-grained information by generating sub-bags with feature distributions similar to the original WSIs. Additionally, we utilize a pseudo-bag generation algorithm to further leverage the abundant and redundant information in WSIs, allowing efficient training in unbalanced-sample multi-classification tasks. Furthermore, we introduce an affinity-based sample selection and curriculum contrastive learning strategy to enhance the stability of model representation learning. Unlike previous approaches, our framework transitions from learning bag-level representations to understanding and exploiting the feature distribution of multi-instance bags. Our method demonstrates significant performance improvements on three datasets, including tumor classification and lymph node metastasis. On average, it achieves a 4.39-point improvement in F1 score compared to the second-best method across the three tasks, underscoring its superior performance.
\end{abstract}

\section{1. Introduction}

\textcolor{clcl2}{Pathology, as a critical component of medical diagnostics, has witnessed significant advancements with the integration of computational techniques.} Computational pathology has significantly enhanced the workflow and diagnostic decision-making processes for pathologists \cite{zhang2022dtfd}. \textcolor{clcl2}{In particular, the advent of computer vision has revolutionized the automated analysis of Whole Slide Images (WSIs), drawing considerable attention from the research community.} Numerous studies have demonstrated that computer-aided diagnostic systems provide substantial reference value for pathologists \cite{lu2021data,lee2022derivation}. 

However, the vast scale of WSIs (billions of pixels) and the extremely high information density (comprising hundreds of thousands of cells and heterogeneous tissues) \cite{juyal2024sc} present challenges for achieving slide-level predictions through end-to-end models. 

\begin{figure}[ht]
    \centering
    \includegraphics[width=\linewidth]{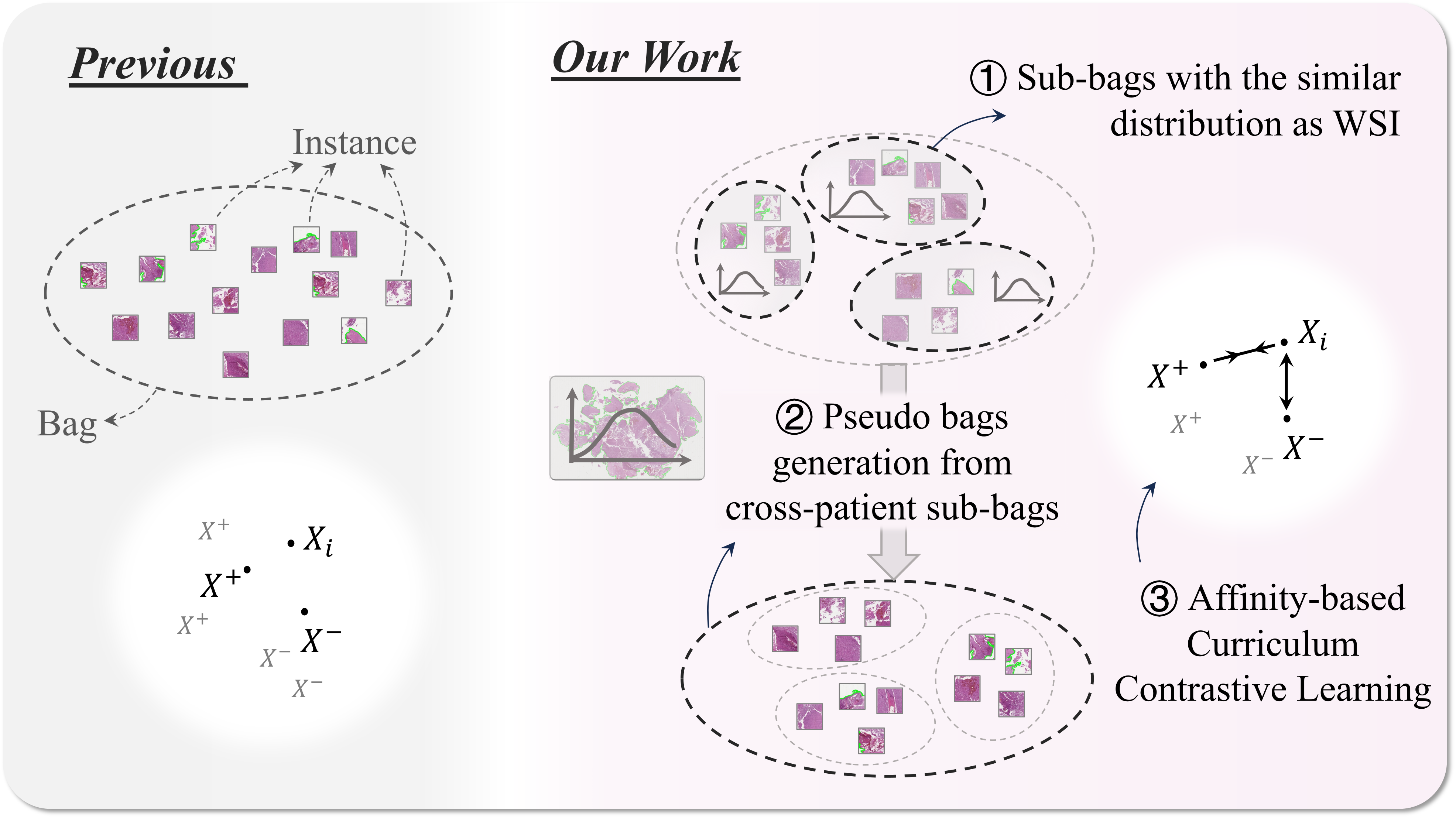}
    \caption{The motivation of our method. Left: Conventional methods aggregate all instances coarsely without considering further efficient utilization of all instances. Right: We propose a cross-patient instance utilization method based on feature distribution and a curriculum contrastive method to improve the expressive ability of the model and alleviate the issue of multi-class imbalance.}
    \label{fig:innovation}
\end{figure}

\textcolor{clcl2}{One promising approach to overcome these challenges is multi-instance learning (MIL), which is particularly effective for WSI classification.} In the context of WSI classification, MIL divides a slide into thousands of smaller image instances and treats this collection as a multi-instance bag (Figure \ref{fig:innovation}). This bag is supervised using slide-level labels rather than instance-level or pixel-level labels. In this approach, if at least one instance within the bag is positive, the entire WSI is labeled as positive; otherwise, it is labeled as negative. \textcolor{clcl2}{Although MIL transforms instance-level classification into slide-level classification, significant challenges remain in complex pathology classification scenarios.}

The difficulty of multi-classification under unbalanced sample conditions is one of the striking challenges. The distribution of categories in pathology image classification is often unbalanced due to differences in disease incidence and degree of progression. For example, certain disease subtypes may be rarer than others, or there may be little data for a particular grade in the pathology grading. This imbalance is not only observed in the overall distribution of the dataset but also within each WSI. For example, only a fraction of malignant regions may contribute to the positive labels in a WSI, making it challenging to detect these regions accurately. This unbalanced category distribution becomes even more problematic in multi-classification tasks, potentially leading to poor model performance for rare categories, even though some rare lesions may be of significant clinical importance.

\textcolor{clcl2}{To address the problem of category imbalance, previous studies have primarily focused on enhancing the representation capabilities of models. This includes the use of pre-trained feature extractors, multi-scale feature fusion, complex instance mining, and curriculum learning approaches (details in section 2).} These methods aim to cope with sample imbalance by improving feature representation or learning strategy. WSI image information has huge redundancy, and previous methods seek to extract salient features while removing redundancy. \textcolor{clcl2}{However, we propose that the redundancy of WSI information can provide an alternative solution to the unbalanced sample multi-classification problem.}

We propose a framework to utilize redundant information in WSIs by reorganizing feature distribution to address multi-classification with unbalanced samples. Pathology images contain complex cellular structures and tissue patterns. By dividing images into sub-bags with similar features, we can better capture and analyze these fine-grained pathological features. Specifically, we construct pseudo-bags across patients using sub-bags, leveraging the rich information in pathology images.

To ensure strong feature representation, we use an affinity-based sample selection method for contrastive learning with adjustable difficulty. This reduces category imbalance and improves model generalization from limited samples. Our contributions are:
\begin{itemize}
\item \textcolor{clcl2}{A two-stage instance bag generation strategy to improve category balance during training. First, generate sub-bags from WSIs to capture feature distributions. Then, combine sub-bags from different patients within the same category to create pseudo-bags, enhancing balanced training and model robustness.}
\item \textcolor{clcl2}{An affinity-based sample selection strategy and a curriculum contrastive learning approach to optimize the feature space.}
\item \textcolor{clcl2}{Significant performance improvement on three datasets, addressing multi-classification imbalance in pathology image classification.}
\end{itemize}

\section{2.Related Work} \label{relatedwork}
\subsection{2.1 Multiple Instance Learning in WSI}
In recent years, MIL has been widely used in WSI analysis \cite{carbonneau2018multiple}. MIL is a weakly supervised framework that organizes data into bags, each containing multiple instances but only one bag-level label, differing from traditional supervised learning where each sample has a corresponding label. The high resolution of WSI makes it difficult to directly input them into deep learning models on existing hardware devices, and manual annotation of WSI images is challenging due to their large size. MIL methods effectively address these challenges in WSI analysis.

Significant progress has been made in instance-level and embedding-level MIL research \cite{li2021dual}. In instance-level MIL, key instances within a bag are selected to represent the entire bag and are assigned a bag-level label. Then, an instance-level classifier is trained, and the final bag-level prediction requires aggregating the prediction results of all instances. The selection of key instances can be achieved using the EM algorithm \cite{hou2016patch} or a histogram-based feature selection algorithm \cite{zhao2020predicting}. 

Other works \cite{chikontwe2020multiple, kanavati2020weakly} iteratively train neural networks to facilitate instance-level classifiers. One such method selects key instances and integrates their classification results into the overall WSI prediction using RNN \cite{campanella2019clinical}.

\textcolor{clcl}{In embedding-based methods, a high-level representation of a bag is learned to build a bag-level classifier. Existing research has proposed various aggregation methods, including clustering \cite{yang2023hamil}, attention mechanisms \cite{ilse2018attention, li2021dual,lu2021data, xiang2023exploring}, transformers \cite{shao2021transmil, yu2023local}, and graph networks \cite{chen2021whole, lee2022derivation}.}

Attention-based methods, such as \cite{ilse2018attention, li2021dual, lu2021data}, use neural networks to predict the weights of each instance and perform weighted averaging of instance features based on these weights. Transformer-based methods \cite{shao2021transmil, yu2023local, fourkioti2023camil} incorporate the query mechanism of transformers into weight prediction, enhancing interaction between instances. Graph-based methods \cite{hou2016patch,li2018graph, lee2022derivation, zheng2022graph} utilize a graph structure to preserve spatial relationships between instances and aggregate information of WSI through graph networks. TEA-Graph \cite{lee2022derivation} further introduces super-patch to effectively reduce node numbers and improve training efficiency.

\begin{figure*}[ht]
    \centering
    \includegraphics[width=0.9\linewidth]{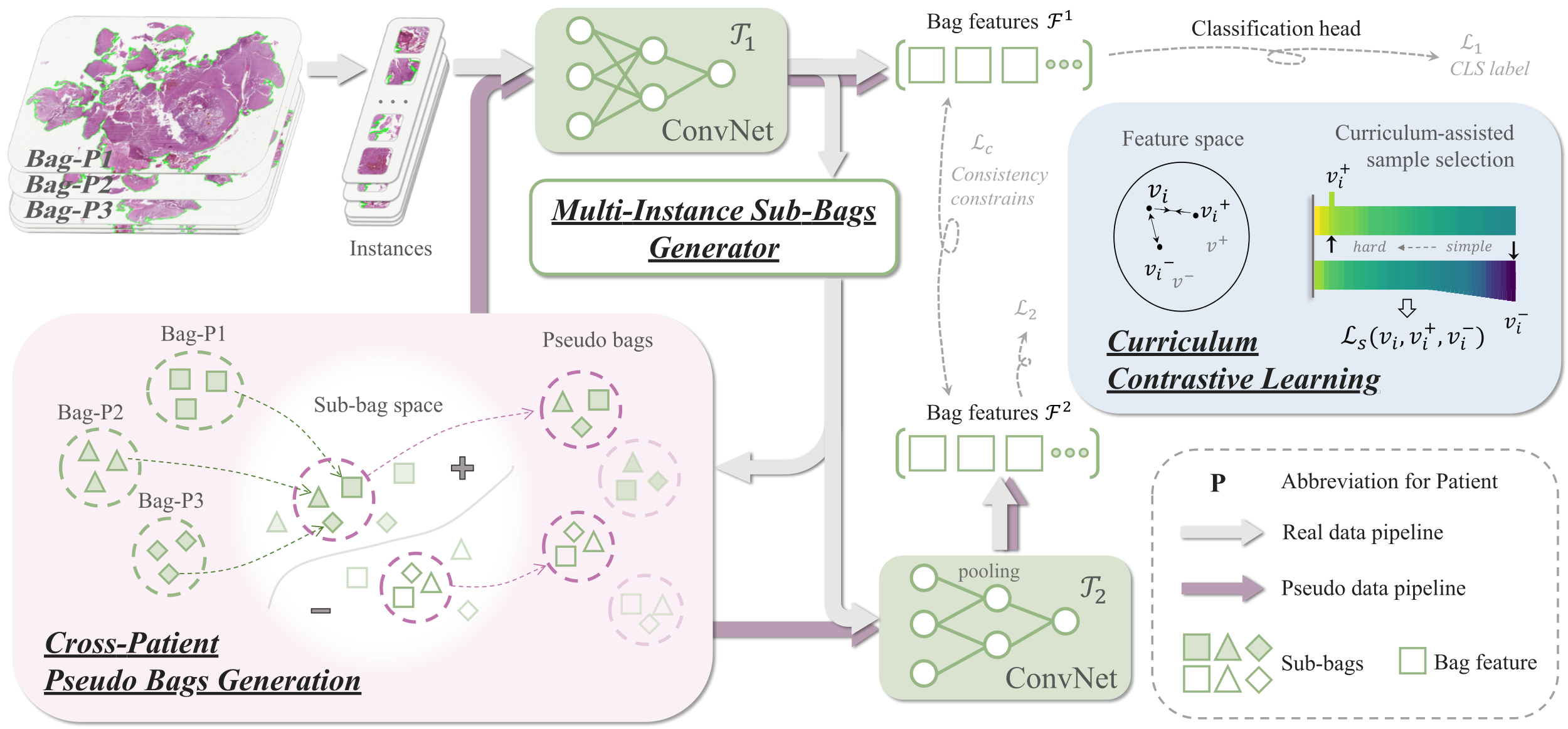}
    \caption{Framework Overview: We process pathological images into multi-instance bags and generate sub-bags with similar feature distributions. These cross-patient sub-bags are combined to create pseudo-bags, leveraging the redundant information in WSIs. An affinity-based sample selection method is employed to achieve curriculum contrastive learning. These techniques collectively improve the model's performance in addressing multi-class imbalance in pathological tasks.}
    \label{fig:method}
\end{figure*}

\subsection{2.2 Representation Learning in WSI}
Learning is the process of transforming raw data into more expressive and interpretable feature representations. In deep learning-based WSI analysis, it is challenging to directly input billion-pixel-level images into the network. Therefore, thousands to tens of thousands of patches need to be extracted for feature extraction. Existing methods \cite{ilse2018attention, lu2021data, zhang2022dtfd,shao2021transmil} typically use pre-trained models on ImageNet, which poses challenges in obtaining high quality feature representations.

To achieve better feature representations, existing research has proposed methods such as feature extractor pre-training, multi-scale feature fusion, complex instance mining, contrastive learning, task-specific fine-tuning, curriculum learning. For pre-training, VAEGAN \cite{zhao2020predicting} trains feature extractors using a combination of variational autoencoders and generative adversarial networks instead of traditional pre-training on classification networks. Other methods \cite{lazard2023giga, li2021dual} introduce self-supervised methods on natural images, considering multiple instance bags from the same WSI as positive samples and bags from different WSIs as negative samples, rather than using masked images \cite{filiot2023scaling} or separating positive and negative regions within the WSI \cite{wang2022scl}.

For multi-scale fusion, many studies \cite{li2021dual, liu2023dsca, huang2023deep} extract instance images at multiple resolutions during preprocessing for subsequent feature embedding and fusion. Another study \cite{thandiackal2022differentiable} proposes an end-to-end scaling method to obtain multi-resolution information, reducing the cost of preprocessing.

Mining complex instances is also a way to improve model representation. HNBG \cite{li2019deep} and MHIM-MIL \cite{xiang2023exploring} \textcolor{clcl}{use instance-level attention mechanisms to filter out simple examples and focus on complex ones}. MHIM-MIL further enhances this approach by using a teacher network of instructors to guide student models in learning complex instance-based classification tasks.

In the research direction of contrastive learning, the SGCL \cite{lin2023sgcl} fully exploits the intrinsic properties of pathology images by utilizing spatial similarity and multi-object priors to learn object differences between different image patches at various fine-grained scales. MuRCL \cite{zhu2023murcl} extends this approach by designing a reinforcement learning strategy to progressively update the selection of the discriminative feature set based on online rewards, guiding WSI-level feature aggregation.

For task-specific fine-tuning, Prompt-MIL \cite{zhang2023prompt} employs trainable visual prompts to fine-tune the feature extractor, reducing the discrepancy between the pre-trained task and the specific task. Another approach \cite{li2023task} introduces the information bottleneck theory, enhancing performance and generalization by maximizing the information bottleneck objective. This approach makes the latent representation vectors more predictive while filtering out irrelevant features.

In summary, our research diverges from the surveyed works in several key aspects. Rather than learning bag-level representations, we delve into understanding and exploiting the feature distribution within multi-instance bags. Specifically, we generate sub-bags with similar feature distributions and utilize these sub-bags to achieve cross-patient pseudo bag generation. Additionally, we propose a curriculum contrastive learning method. This comprehensive approach results in a more nuanced and detailed representation of the data. Workflow of the proposed framework is shown in Figure \ref{fig:method}.

\section{3.Method}
\subsection{3.1 Feature Distribution-Guided Sub-bag and Pseudo-Bag Generation}
Consider a WSI collection $X_{i} \in\{X_{1}, \ldots, X_{P}\}$ containing $P$ patients with the corresponding label $Y_{i} \in\{Y_{1}, \ldots, Y_{P}\}$. Each pathology image is tiled and extracting non-overlapping $N$ foreground patches $\{X_{i j}\}_{j=1}^{N}$ by a tissue segmentation method. These patches serve as instances in a weakly supervised task, where instance-level labels are not provided. A feature extraction network is applied to embed all patches $\{X_{i j}\}_{j=1}^{N}$ to a bag of feature vectors $\{f_{i j}\}_{j=1}^{N}$. Following that, an aggregation methods or feature selection algorithm is employed to obtain a bag feature vector, which is then used as the input to a classifier for predicting the patient-level label $Y_i$.

\textcolor{clcl}{We extend the label prediction task to predict the joint distribution of labels $Y$ and latent variables $O$, where the latent variables $O$ describe the distribution of all regions in the pathological images. We assume that the latent variables $O$ follow the distribution $Q(O)$. Using variational inference, we seek the expression for the variational distribution $Q(O_i)$:}
\begin{equation}
\begin{aligned}
\sum_{i} \ln P(Y_{i} \mid X_{i}, \theta) &\geq \sum_{i} \sum_{O_{i}} Q(O_{i}) \\
&\quad \ln (\frac{P(Y_{i}, O_{i} \mid X_{i}, \theta)}{Q(O_{i})}),
\end{aligned}
\end{equation}
here, $\theta$ represents the parameters of learnable model. The goal of variational inference here is to maximize the above likelihood function. Specifically, $\sum_{i} \ln P(Y_{i} \mid X_{i}, \theta)$ is equivalent to its lower bound if and only if $Q(O_{i}) = P(O_{i} \mid Y_{i}, X_{i}, \theta)$. Therefore, we can use $Q(O)$ to approximate the posterior distribution $P(O_i \mid Y_i, X_i, \theta)$.

In practical applications, directly calculating and optimizing the Evidence Lower Bound (ELBO) can be quite complex. Therefore, we propose a simplified method: further approximating $P(O_i \mid Y_i, X_i, \theta)$ with $P(O_i \mid X_i, \theta)$. \textcolor{clcl}{Since the neural network implicitly learns the mapping from $X$ to $Y$ during the training process, using $P(O_i \mid X_i, \theta)$ as an approximation for $P(O_i \mid Y_i, X_i, \theta)$ is reasonable. }

Based on this approximation, we can express the joint probability $ P(Y, O \mid X, \theta) $ as follows:
\begin{equation}
P(Y, O \mid X, \theta)=P(Y \mid X, O, \theta) P(O \mid X, \theta) ,
\end{equation}
where $P(Y \mid X, O, \theta)$ and $P(O \mid X, \theta)$ constitute the differentiable function $\mathcal{T}_{1}: \mathbb{R}^{N \times d} \rightarrow$ $\mathbb{R}^{1 \times c}$, where $N$ is the number of the instances in a whole slide image $X$, $d$ is the embedding dimension. The $\mathcal{T}_{1}$ first generates feature distribution $Q(O_{i})=\{\beta_{i j}\}_{j=1}^{N}$ for each pathological image and then aggregates all remapped instance features  into $v_{i} \in \mathbb{R}^{1 \times v}$ using $Q(O_{i})$ for final classification. We formulate the learning objective of this part as follows: 
\begin{equation}
\mathcal{L}_1=\underset{X_i, Y_i \sim X, Y}{\mathbb{E}} \mathcal{H}(\mathcal{T}_1(\{f_{i j}\}_{j=1}^N), Y_i),
\end{equation}
where $\mathcal{H}(\cdot)$ is the cross entropy loss function.

A learnable function $\mathcal{T}_2: \mathbb{R}^{S \times \frac{N}{S} \times d} \rightarrow \mathbb{R}^{S \times v} \rightarrow \mathbb{R}^{1 \times c}$ is employed to derive classification vectors based on multiple instance sub-bags. According to the task-specific feature distribution $P(O \mid X, \theta)$, the sub-bags are sampled from multiple instance bags and satisfies the understanding of the overall feature distribution of pathological images by $\mathcal{T}_1$ at different training stages.

In detail, we sort the $\{\beta_{i j}\}_{j=1}^{N}$ and denote the sorted result as $\{\tilde{\beta}_{ij}\}_{j=1}^{N}$, followed by dividing it into $\{\{\tilde{\beta}_{i j}\}_{j=k, s=S}^{N}\}_{k=1}^{k=S}$ according to a certain step $S$ (equal to the number of the sub-bags), which operation is abstracted as a multi-instance sub-bagging function $\mathcal{S}$. Corresponding sorted instance feature set is denoted as $\mathcal{A} = \mathcal{S}(\{\tilde{f}_{ij}\}_{j=1}^{N}) = \{\{\tilde{f}_{ij}\}_{j=k, s=S}^{N}\}_{k=1}^{k=S}$. Through learnable $\mathcal{T}_2$, each sub-bag-level aggregated feature is added to a new set, e.g., $\{v_{i}^{1}, \ldots, v_{i}^{S}\}$, and then yielding the slide-level prediction results through average pooling.

It would be optimized with supervised learning objective as follows:
\begin{equation}    
\mathcal{L}_2=\underset{X_i, Y_i \sim(X, Y)}{\mathbb{E}} \mathcal{H}(\mathcal{T}_2(\mathcal{A}), Y_i) .
\end{equation}

An important implementation in this section is generating pseudo bags by selecting sub-bags from different patients, ensuring each pseudo bag $G_{i}=\{G_{i j}\}_{j=1}^{N}$ consists of $S$ sub-bags with $Y_{i}=c$, \textcolor{clcl}{where $c$ is one of the $C$ categories}. These pseudo bags are cross-patient in nature. We use pseudo bags as unlabeled data and constrain the consistency of their predictions in $\mathcal{T}_1$ and $\mathcal{T}_2$.
\begin{equation}
\mathcal{L}_{c1}=\underset{G_i \sim G}{\mathbb{E}} \mathcal{H}(\mathcal{T}_1(G_i), \mathcal{T}_2(\mathcal{S}(G_i))).
\end{equation}

Additionally, we define $\mathcal{L}_{c2}$ using the original multi-instance bags $\{f_{i j}\}_{j=1}^{N}$ to ensure the consistency of predictions for a bag in $\mathcal{T}_1$ and $\mathcal{T}_2$. Finally, we combine these two consistency losses into a single loss $\mathcal{L}_{c} = \mathcal{L}_{c1} + \mathcal{L}_{c2}$.

\textcolor{clcl}{By generating additional samples from another class during each training iteration, this method gradually balances the distribution of instances across all classes. Specifically, in a classification task with $C$ classes, for each current sample, an additional sample from another class is generated for training. After one iteration, the total number of samples for class $c$, $N_{c}^{\prime}$, is calculated as follows:}
\begin{equation}
N_{c}^{\prime} = N_{c} + \tilde{N}_{c} = N_{c} + \sum_{d \neq c} \frac{N_{d}}{C-1},
\end{equation}
\textcolor{clcl}{where $N_{c}$ represents the initial number of samples for class $c$, $\tilde{N}_{c}$ represents the additional samples generated for class $c$ through the dynamic sample generation method.}

\textcolor{clcl}{This method continuously adjusts and generates new samples during training, leading to a more balanced number of samples across all classes, thereby improving the performance of the classification model on imbalanced datasets.}

\subsection{3.2 Affinity-Based Curriculum Contrastive Learning}

\textcolor{clcl}{To ensure high-performance feature embedding from bags to pseudo-bags, we propose a curriculum contrastive learning strategy.} \textcolor{clcl}{While curriculum learning can stabilize model training, selecting appropriate positive and negative samples throughout the training process remains a challenge. We address this by using a category affinity index to select positives and semi-hard negatives, ensuring that anchor-negative pairs are farther apart than anchor-positive pairs.} This progressively increases the curriculum difficulty and improves model optimization. The algorithm is shown in Algorithm \ref{alg:curriculum}.

\begin{algorithm}[!ht]
\small
\captionsetup{font=small}
\caption{Affinity-based Curriculum Contrastive Learning}
\begin{algorithmic}[1]
\Require Dataset $X$, labels $Y$, momentum coefficient $m$, margin $\alpha$, max epochs $E$
\State Initialize dictionary $D_b$ with embeddings from each case in the first epoch
\For{epoch $e = 1$ to $E$}
    \For{each case $X_i$ in dataset}
        \State $v_i \gets$ Compute embedding for case $X_i$
        \State $D_{bi} \gets m \cdot D_{bi} + (1 - m) \cdot v_i$ \Comment{Update dictionary}    
    \EndFor
    
    \For{each case $X_i$ in dataset}
        \State $I_{c} \gets$ Compute cosine similarities between $v_i$ and $D_b$
        
        \State $I_{\tau} \gets$ Select a threshold for positive sample selection
        
        \State $v_i^+ \gets$ Select from $I_{Y_i}$ with similarity greater than $I_{\tau}$ \Comment{Positive sample}
        
        \State $I_{\neg Y_i} \gets$Filter embeddings in $I_{\neg Y_i}$ with similarity $\leq I_{\tau}$
        
        \State $k \gets 1 - (\frac{e}{E})^2$ \Comment{Adjust curriculum difficulty}
        
        \State $v_i^- \gets$ Select semi-hard negative based on $k$ from $I_{\neg Y_i}$ \Comment{Negative sample}

        \State Construct T triplets $\{(v_{i}, v_{i}^{t+}, v_{i}^{t-})\}$
        
        \For{each triplet $t$ in $T$ triplets}
            \State $\mathcal{L}_s$$\gets$$\mathcal{L}_s$+$\max(0, \operatorname{sim}(v_i, v_i^{t-})$-$\operatorname{sim}(v_i, v_i^{t+})$+$\alpha)$
        \EndFor
    \EndFor
    
    \State Update model parameters using $\mathcal{L}_s$
\EndFor
\end{algorithmic}
\label{alg:curriculum}
\end{algorithm}

First, we maintain a dictionary $D_{b}$ where each key is a unique case ID and its value is the feature embedding $v_i$ for that case. During each epoch, we update the dictionary values using momentum to ensure consistent bag-level representations \cite{he2020momentum}. Specifically, for each case $i$, the embedding $v_i$ is updated as follows:
\begin{equation}
D_{bi} \leftarrow m D_{bi} + (1 - m) v_i,
\end{equation}
where $m \in [0, 1)$ is the momentum coefficient.

We calculate the cosine similarity between each anchor embedding $v_{i}$ and the embeddings in $D_{b}$ to obtain the category affinity index $I_{c \in \{Y_{i}, \neg Y_{i}\}} \in \mathbb{R}^{|D_{b}^c|}$, where $|D_{b}^c|$ is the number of samples labeled with class $c$. Each value in $I_{c}$ represents the similarity between the momentum-updated $D_{bi}$ and the anchor embedding $v_{i}$. The similarity is measured by cosine similarity as follows:
\begin{equation}
\operatorname{sim}(v_{i}, D_{bi}) = \frac{v_{i} \cdot D_{bi}^{\top}}{\|v_{i}\|_{2} \cdot \|D_{bi}\|_{2}}.
\end{equation}

We randomly select a value $I_{\tau} \in I_{c \in \{Y_{i}\}}$ with the top K in similarity as the threshold and determine an embedding corresponding to $I_{\tau}$ as $v_{i}^{+}$. Then, we delimit the selection space in $I_{c \in \{\neg Y_{i}\}}$ according to $I_{\tau}$, where the selection space is $I_{\neg Y_{i}} \leq I_{\tau}$.

A hyper-parameter $k \in [0, 1]$ is introduced to measure the curriculum difficulty. Specifically, the most confusing but semi-hard anchor-negative pair compared to the anchor-positive pair can be determined by $k = 0$. Correspondingly, the anchor-negative pair that ensures the maximum $\operatorname{sim}(v_{i}, v_{i}^{-})$ is selected through $k = 1$. These two cases are described as follows:
\begin{equation}
\begin{aligned}
& k=0 \rightarrow \underset{v_{i}^{-}}{\operatorname{argmin}} \left(\operatorname{sim}(v_{i}, v_{i}^{+}) - \operatorname{sim}(v_{i}, v_{i}^{-})\right), \\
& k=1 \rightarrow \underset{v_{i}^{-}}{\operatorname{argmax}} \left(\operatorname{sim}(v_{i}, v_{i}^{+}) - \operatorname{sim}(v_{i}, v_{i}^{-})\right).
\end{aligned}
\end{equation}

Relaxing curriculum difficulty stabilizes gradients in early training. As training progresses, simple objectives become insufficient, so increasing difficulty helps the model escape local optima. We adjust the difficulty parameter $k$ using $k = 1 - \left(\frac{\text{epoch}}{\text{max epoch}}\right)^2$ to ensure a gradual increase in difficulty.

We determine $v_{i}^{+}$ and $v_{i}^{-}$ through $I_{c \in \{Y_{i} \rightarrow Y_{i}\}}$ and $k_{i}$ in each iteration. To speed up the construction of positive and negative pairs for the model, $T$ triplets are constructed in each iteration according to the strategy mentioned above, e.g., $\left\{(v_{i}, v_{i}^{1+}, v_{i}^{1-}), \ldots, (v_{i}, v_{i}^{T+}, v_{i}^{T-})\right\}$.

We minimize the similarity between the same classes and maximize the similarity between different classes according to the curriculum. The objective function is formulated as follows:
\begin{equation}
\mathcal{L}_s = \sum_{i=1}^{P} \sum_{t=1}^{T} \max \left(0, \operatorname{sim}(v_i, v_i^{t-}) - \operatorname{sim}(v_i, v_i^{t+}) + \alpha \right),
\end{equation}
where $P$ is the size of the WSI collection, $\operatorname{sim}(v_i, v_i^{t+})$ represents the similarity between sample $v_i$ and the positive sample $v_i^{t+}$, $\operatorname{sim}(v_i, v_i^{t-})$ is analogous.

\section{4.Experiments}
\subsection{4.1 Datasets and Implementation Details}

The Cancer Genome Atlas (TCGA) Kidney dataset includes 887 WSIs in SVS format across three subtypes: 500 KIRC, 273 KIRP, and 114 KICH slides. The Liver Cancer dataset, available upon request, contains 670 HCC, 206 CHC, and 92 ICC WSIs. The CAMELYON17 \cite{bejnordi2017diagnostic} comprises 500 publicly available WSIs for detecting breast cancer metastases, with 5 WSIs per patient. It includes 87 Macro, 59 Micro, 36 ITC, and 318 Normal instances. We combined Macro, Micro, and ITC into a Positive class, while Normal remains tumor-negative. 

We randomly selected 1/5 of the dataset as the test set, using the rest for training and validation. The training-to-validation ratio was 4:1, and all results are reported as the average of five experiments.

All methods are based on the same data preprocessing tools, which follows the CLAM \cite{lu2021data}. Our experiments utilized two model structures: the first structure, $\mathcal{T}_1$, is identical to the CLAM, while the second structure, $\mathcal{T}_2$, adds a pooling module on top of the CLAM model. During training, a weighted average loss function was used, specifically incorporating $\mathcal{L}_1$, $\mathcal{L}_2$, $\mathcal{L}_s$, and $\mathcal{L}_c$. The experiment setup includes a maximum of 100 epochs, with early stopping enabled and a patience value of 20. The optimizer used is Adam, with a learning rate of 1e-4 and a regularization parameter of 2e-5. The batch size for all experiments is set to 1. Model performance evaluation metrics follow the standards outlined in \cite{yu2020lymph}, including F1 score, area under the receiver operating characteristic curve (AUC), accuracy (ACC), Matthew’s correlation coefficient (MCC), sensitivity (SENS), specificity (SPEC), positive predictive value (PPV), and negative predictive value (NPV). All experiments were conducted on a system running Ubuntu 18.04.6 LTS, equipped with an NVIDIA V100 32G GPU, using CUDA version 11.7, Python version 3.7.7, and PyTorch version 1.12.1+cu113.

\subsection{4.2 Performance Comparison with Existing Works}
We demonstrated the experimental results of the proposed method on the three datasets, and compared them with SOTA methods including standard DSMIL\cite{li2021dual}, TransMIL\cite{shao2021transmil}, DTFD\cite{zhang2022dtfd}, and CLAM\cite{lu2021data}. 

The TCGA Kidney and Liver Cancer datasets are used for tumor classification tasks, while the Camelyon17 dataset is used for lymph node metastasis classification tasks.

Table \ref{tab:sota} highlights the challenge of class imbalance in pathology image research. Advanced baseline methods show high AUC scores but lower class average F1 scores.

The DTFD method, based on standard MIL, randomly splits the multi-instance bag into multiple sub-bags and completes classification by separately selecting and integrating instance features. In tumor classification tasks with class imbalance, the AUC is much higher than the F1 score, indicating that DTFD may not effectively extract key instances to represent the class within the sub-bags, particularly for minority classes.

Methods based on bag feature integration, such as CLAM, TransMIL, and DSMIL, perform more consistently across the three tasks. Our method achieves the highest scores in metrics that emphasize handling imbalanced scenarios, such as F1 score (default: macro), AUC, and MCC.

\begin{table}[ht]
\centering
\tiny
\caption{Classification results on three datasets.}
\begin{tabular}{ccccccccc}\hline
\multicolumn{9}{c}{(a) TCGA Kidney} \\ \hline
\multicolumn{1}{c|}{Methods}  & F1             & AUC            & ACC            & MCC           & SEN            & SPE            & PPV            & NPV            \\ \hline
\multicolumn{1}{c|}{DSMIL}    & 66.24          & 95.47          & 80.00          & 58.09         & 66.45          & 87.79          & 77.60          & 89.48          \\
\multicolumn{1}{c|}{TransMIL} & 73.78          & 95.27          & 84.04          & 68.16         & 72.55          & 90.77          & 86.64          & 91.77          \\
\multicolumn{1}{c|}{CLAM}     & 64.75          & 94.84          & 79.55          & 55.81         & 65.68          & 87.65          & 70.57          & 89.33          \\
\multicolumn{1}{c|}{DTFD}     & 57.76          & 94.45          & 82.13          & 50.62         & 62.07          & 89.26          & 54.22          & 91.81          \\
\multicolumn{1}{c|}{Ours}     & \textbf{84.21} & \textbf{97.47} & \textbf{88.88} & \textbf{79.28}& \textbf{82.40} & \textbf{93.84} & \textbf{89.46} & \textbf{94.11} \\ \hline

\multicolumn{9}{c}{(b) Liver Cancer}  \\ \hline
\multicolumn{1}{c|}{Methods}  & F1             & AUC            & ACC            & MCC           & SEN            & SPE            & PPV            & NPV            \\ \hline
\multicolumn{1}{c|}{DSMIL}    & 87.42          & \textbf{99.13} & 92.64          & 84.43         & \textbf{92.70} & 96.99          & 85.87          & 94.59          \\
\multicolumn{1}{c|}{TransMIL} & 70.24          & 94.32          & 84.56          & 62.41         & 66.88          & 87.19          & 85.03          & 91.32          \\
\multicolumn{1}{c|}{CLAM}     & 84.24          & 98.92          & 92.12          & 80.58         & 85.10          & 94.77          & \textbf{87.48} & \textbf{95.44} \\
\multicolumn{1}{c|}{DTFD}     & 32.94          & 92.17          & 56.27          & 18.97         & 41.80          & 72.64          & 37.67          & 71.23          \\
\multicolumn{1}{c|}{Ours}     & \textbf{88.95} & 98.78          & \textbf{93.78} & \textbf{85.54}& 92.47          & \textbf{97.07} & 86.94          & 95.15          \\ \hline

\multicolumn{9}{c}{(c) Camelyon 17}  \\ \hline
\multicolumn{1}{c|}{Methods}  & F1             & AUC            & ACC            & MCC            & SEN            & SPE            & PPV            & NPV            \\ \hline
\multicolumn{1}{c|}{DSMIL}    & 74.80          & 80.14          & 80.00          & 50.87          & 76.86          & 76.86          & 74.14          & 74.14          \\
\multicolumn{1}{c|}{TransMIL} & 77.89          & 86.54          & 82.20          & 57.27          & 80.88          & 80.88          & 76.59          & 76.59          \\
\multicolumn{1}{c|}{CLAM}     & 72.73          & 87.02          & 76.00          & 51.19          & 79.36          & 79.36          & 72.31          & 72.31          \\
\multicolumn{1}{c|}{DTFD}     & 80.04          & 80.92          & \textbf{85.80} & 61.01          & 78.97          & 78.97          & \textbf{82.29} & \textbf{82.29} \\
\multicolumn{1}{c|}{Ours}     & \textbf{80.86} & 84.63          & 85.20          & \textbf{62.66}& \textbf{82.00} & \textbf{82.00} & 80.82          & 80.82          \\ \hline
\end{tabular}
\label{tab:sota}
\end{table}

\subsection{4.3 Heatmap and Class Imbalance Visualization}
\begin{figure}[ht]
    \centering
    \includegraphics[width=\linewidth]{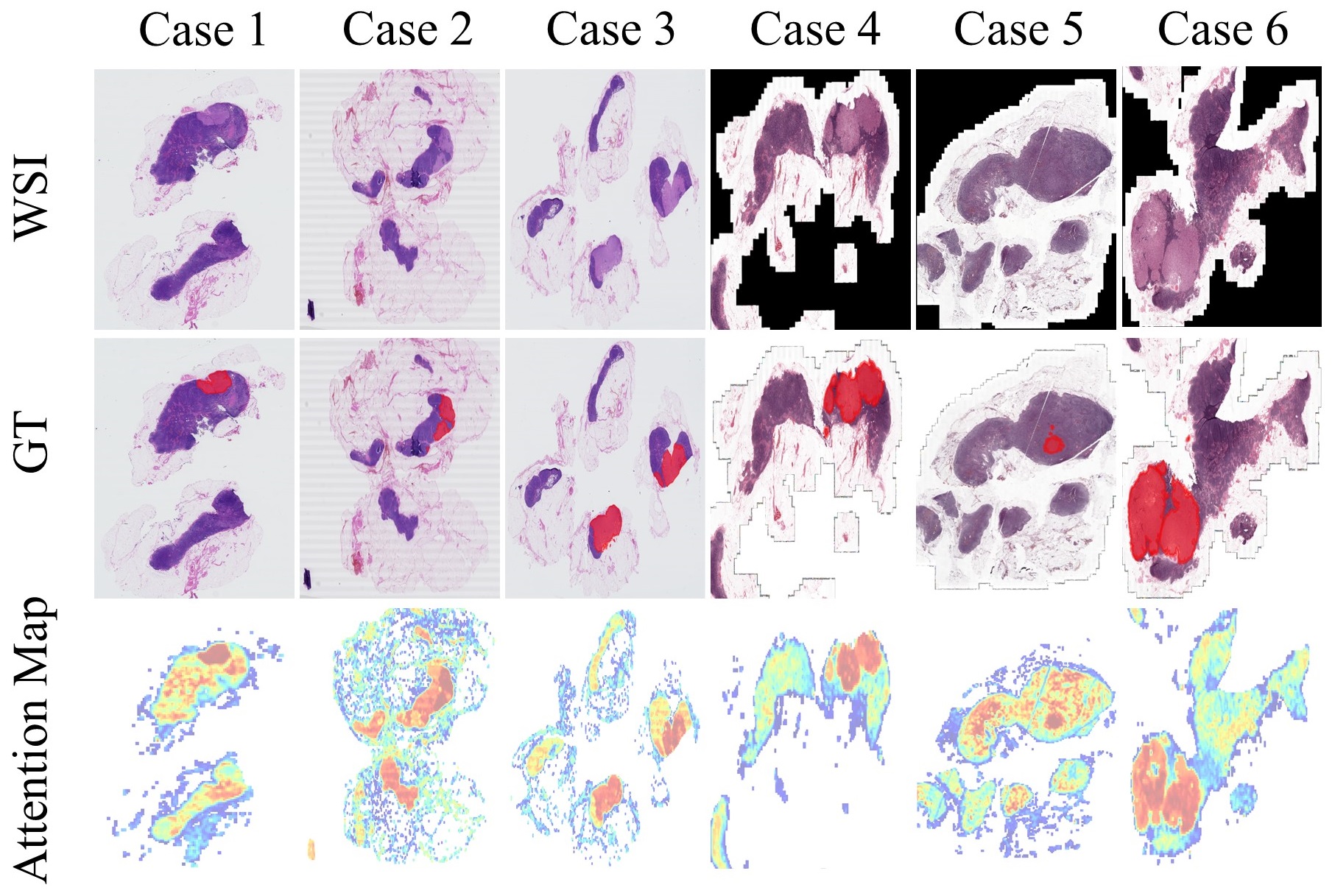}
    \caption{Tumor localization in the whole slide images.}
    \label{fig:heatmap}
\end{figure}

For the Camelyon 17, we visualized the model using attention scores to obtain the tumor locations. We normalized the attention scores and mapped them to a color space to draw heatmaps for each WSI based on the spatial location of the instances. We selected five WSIs from the Camelyon 17 for demonstration, as shown in Figure \ref{fig:heatmap}. The tumor areas are highlighted in red in the second row of the figure, and the heatmaps in the third row show clear high attention, illustrating the model's potential for tumor localization.

For TCGA Kidney, a significant class imbalance exists with a ratio of 1:2.4:4.4. This imbalance may lead to suboptimal embedding of features by the model, making it difficult to adapt to new data. We trained twice for each method and projected their feature embedding into a two-dimensional space.

\begin{figure}[ht]
    \centering
    \includegraphics[width=\linewidth]{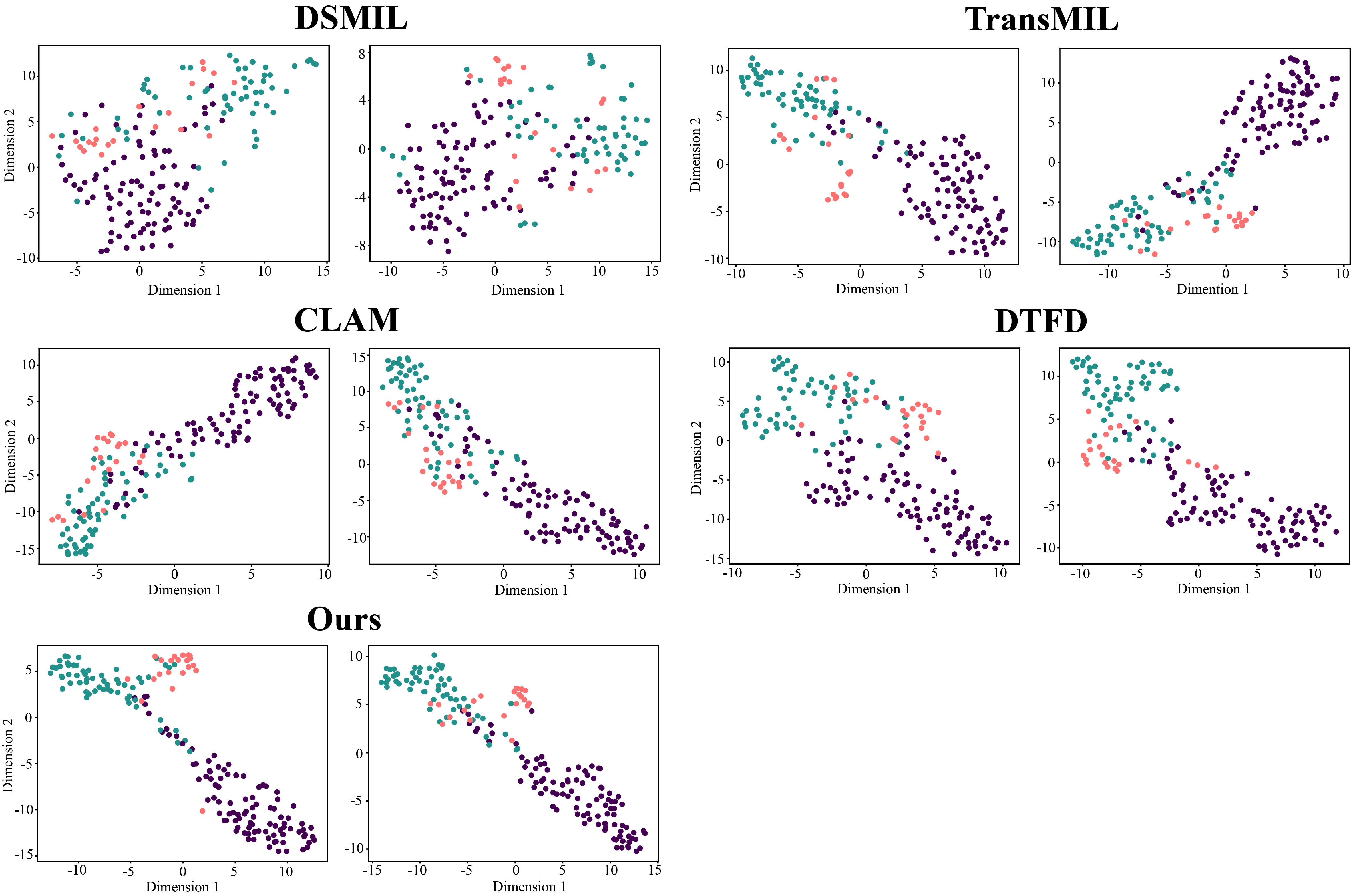}
    \caption{Visualization of feature distribution on the TCGA Kidney using t-SNE.}
    \label{fig:tsne}
\end{figure}

In Figure \ref{fig:tsne}, the class imbalance is evident. The DTFD, based on group instance selection, did not yield better spatial representation than the Top K-based instance selection (MIL). Among embedding-based methods, our approach not only exhibited better classification boundaries between multiple sample classes but also effectively grouped small samples into more compact clusters. 

\subsection{4.4 Multi-Class Imbalance}
This study thoroughly investigated the challenge of multi-category imbalance in pathology image classification, focusing on TCGA kidney with original sample ratios of 1:2.4:4.4 across three categories. To systematically evaluate the performance of our proposed method under various imbalance conditions, we generated several imbalanced datasets with the following ratios: 1:1:1, 1:1:2, 1:1:4, 1:2:2, and 1:2:3. These ratios encompass a wide range of scenarios, from perfectly balanced to highly imbalanced. Entropy is a measure of uncertainty or disorder, and for a set of categories with proportions $ p_1, p_2, \ldots, p_n $, it is defined as:$ -\sum_{i=1}^{n} p_i \log(p_i) $, where $ p_i $ is the proportion of samples in category $i$. This formula helps capture the degree of imbalance, with higher entropy indicating a more balanced distribution and lower entropy indicating greater imbalance, as detailed in Table \ref{tab:imblance}.

\begin{table}[ht]
\centering
\tiny
\caption{Results under various class imbalance configurations. The entropy values are calculated to quantify the degree of imbalance. The "Cases" column indicates the relative number of samples.}
\begin{tabular}{cccccc}
\hline
\begin{tabular}[c]{@{}c@{}}Imbalance \\ Ratio\end{tabular} & Entropy & Cases & F1 Score & AUC   & ACC   \\ \hline
1:1:1                                                      & 1.5850  & 3x    & 79.62    & 96.42 & 84.94 \\
1:1:2                                                      & 1.5000  & 4x    & 81.17    & 96.34 & 85.62 \\
1:1:4                                                      & 1.2516  & 6x    & 81.06    & 96.90 & 86.40 \\
1:2:2                                                      & 1.5219  & 5x    & 77.04    & 95.95 & 84.61 \\
1:2:3                                                      & 1.4591  & 6x    & 81.41    & 96.74 & 87.42 \\
\hline
\end{tabular}
\label{tab:imblance}
\end{table}

From Table \ref{tab:imblance}, we observed that our method maintains stable performance even in the presence of significant data imbalance. For instance, under the 1:1:4 imbalance configuration, the entropy value drops to 1.2516, indicating a high degree of data imbalance. Nevertheless, the model's F1 Score, AUC, and ACC reached 81.06, 96.90, and 86.40, respectively, which are comparable to, and even slightly better than, the performance under the balanced configuration (1:1:1). This demonstrates that our method exhibits notable robustness in the face of data imbalance, maintaining stable performance across different levels of imbalance configurations.

\subsection{4.5 Ablation Study}
We identified three innovative points: C1, the consistency of predictions for a bag in $\mathcal{T}_1$ and $\mathcal{T}_2$.; C2, the use of curriculum contrastive learning; and C3, the generation of pseudo-bags using cross-patient sub-bags. For the ablation study, we removed each innovative point (C1, C2, and C3) separately and conducted experiments to assess their impact on the model's performance.

\begin{table}[ht]
\tiny
\centering
\caption{Ablation study on the components of our method. C1, consistency of predictions for a bag in $\mathcal{T}_1$ and $\mathcal{T}_2$; C2, curriculum contrastive learning; and C3, cross-patient pseudo-bags.}
\begin{tabular}{ccccccc} \hline
Method & C1 & C2 & C3 & F1 Score & AUC & ACC   \\ \hline
Ours   &    &    &    & 84.21      & 97.47 & 88.88   \\
M1     & ×  &    &    & 81.57      & 96.80 & 86.29   \\
M2     &    & ×  &    & 81.62      & 96.71 & 86.85   \\
M3     &    &    & ×  & 80.23      & 96.20 & 86.29   \\
M4     & ×  & ×  &    & 75.84      & 96.30 & 84.94   \\
M5     & ×  & ×  & ×  & 73.94      & 96.35 & 83.82   \\ \hline
\end{tabular}
\label{tab:abla}
\end{table}

\begin{table}[ht]
\tiny
\centering
\caption{Ablation study on the curriculum function and the number of sub-bags. The curriculum functions include Exponential (E), Random (R), Linear (L), and the default Smooth (S) function.}
\setlength{\tabcolsep}{1.5mm}{
\begin{tabular}{cccccccccc}
\hline
\multirow{2}{*}{Ablation} & \multicolumn{4}{c}{Curriculum Function} & \multicolumn{5}{c}{Num of Sub-bags}   \\ \cline{2-10} 
                                & E  & R & L & S & 3     & 5     & 7     & 9     & 11    \\ \hline
F1 Score                        & 69.91        & 79.14  & 82.31  & 84.21  & 75.26 & 80.61 & 78.08 & 83.80 & 84.21 \\
AUC                             & 96.19        & 7.02   & 96.35  & 97.47  & 96.62 & 97.21 & 97.45 & 97.66 & 97.47 \\
ACC                             & 83.15        & 87.64  & 87.41  & 88.88  & 85.51 & 87.53 & 86.74 & 88.88 & 88.88 \\ \hline
\end{tabular}}
\label{tab:abla_two}
\end{table}

The Table \ref{tab:abla} showed that removing any of the innovative points led to a decrease in model performance. Notably, simultaneous removal of C1 and C2 (M4) or removal of all features (M5) resulted in a significant performance decline. 

Additionally, Table \ref{tab:abla_two} explores the effect of varing the number of sub-bags on model performance. Results indicate that increasing the number of sub-bags initially enhances F1 Score (macro), AUC, and ACC, stabilizing after reaching optimal levels. This suggests that while increasing sub-bags can improve performance, the marginal benefit diminishes beyond a certain threshold.

In Table \ref{tab:abla_two}, we compare several different curriculum learning difficulty scheduling functions. Our method uses the smooth function as the default setting, and we compare it with three other functions: exponential (E), random (R), and linear (L). The linear function is defined as $ h = 1 - \frac{\text{epoch}}{\text{max epoch}} $, the exponential function is defined as $ h = \exp\left(-\frac{\text{epoch}}{\text{max epoch}}\right) $, and the random function is defined as $ h = \text{random}(0, 1) $. The experimental results show that using the smooth function achieves the best performance in terms of convergence speed and final classification accuracy. 

\section{5.Conclusion}
In this paper, we address the issue of class imbalance in pathological image classification tasks and leverage the rich and sometimes redundant information in pathological images to enhance the model's representational capacity. Specifically, our approach explores three main aspects. First, we propose obtaining sub-bags with similar feature distributions from the original multi-instance bags, and these sub-bags from different patients can be used to construct pseudo-bags, thereby making more efficient use of the information in pathological images. Finally, we investigate sample selection based on affinity to achieve curriculum contrastive learning. These explorations have been thoroughly validated on three datasets, demonstrating that our approach can effectively improve the imbalanced multiclassification of whole slide images.

\bibliography{aaai25}

\begin{thebibliography}{48}
\providecommand{\natexlab}[1]{#1}

\bibitem[{Bejnordi et~al.(2017)Bejnordi, Veta, Van~Diest, Van~Ginneken, Karssemeijer, Litjens, Van Der~Laak, Hermsen, Manson, Balkenhol et~al.}]{bejnordi2017diagnostic}
Bejnordi, B.~E.; Veta, M.; Van~Diest, P.~J.; Van~Ginneken, B.; Karssemeijer, N.; Litjens, G.; Van Der~Laak, J.~A.; Hermsen, M.; Manson, Q.~F.; Balkenhol, M.; et~al. 2017.
\newblock Diagnostic assessment of deep learning algorithms for detection of lymph node metastases in women with breast cancer.
\newblock \emph{Jama}, 318(22): 2199--2210.

\bibitem[{Campanella et~al.(2019)Campanella, Hanna, Geneslaw, Miraflor, Werneck Krauss~Silva, Busam, Brogi, Reuter, Klimstra, and Fuchs}]{campanella2019clinical}
Campanella, G.; Hanna, M.~G.; Geneslaw, L.; Miraflor, A.; Werneck Krauss~Silva, V.; Busam, K.~J.; Brogi, E.; Reuter, V.~E.; Klimstra, D.~S.; and Fuchs, T.~J. 2019.
\newblock Clinical-grade computational pathology using weakly supervised deep learning on whole slide images.
\newblock \emph{Nature medicine}, 25(8): 1301--1309.

\bibitem[{Carbonneau et~al.(2018)Carbonneau, Cheplygina, Granger, and Gagnon}]{carbonneau2018multiple}
Carbonneau, M.-A.; Cheplygina, V.; Granger, E.; and Gagnon, G. 2018.
\newblock Multiple instance learning: A survey of problem characteristics and applications.
\newblock \emph{Pattern Recognition}, 77: 329--353.

\bibitem[{Chen et~al.(2021)Chen, Lu, Shaban, Chen, Chen, Williamson, and Mahmood}]{chen2021whole}
Chen, R.~J.; Lu, M.~Y.; Shaban, M.; Chen, C.; Chen, T.~Y.; Williamson, D.~F.; and Mahmood, F. 2021.
\newblock Whole slide images are 2d point clouds: Context-aware survival prediction using patch-based graph convolutional networks.
\newblock In \emph{Medical Image Computing and Computer Assisted Intervention--MICCAI 2021: 24th International Conference, Strasbourg, France, September 27--October 1, 2021, Proceedings, Part VIII 24}, 339--349. Springer.

\bibitem[{Chikontwe et~al.(2020)Chikontwe, Kim, Nam, Go, and Park}]{chikontwe2020multiple}
Chikontwe, P.; Kim, M.; Nam, S.~J.; Go, H.; and Park, S.~H. 2020.
\newblock Multiple instance learning with center embeddings for histopathology classification.
\newblock In \emph{Medical Image Computing and Computer Assisted Intervention--MICCAI 2020: 23rd International Conference, Lima, Peru, October 4--8, 2020, Proceedings, Part V 23}, 519--528. Springer.

\bibitem[{Clancey(1979)}]{c:79}
Clancey, W.~J. 1979.
\newblock \emph{{Transfer of Rule-Based Expertise through a Tutorial Dialogue}}.
\newblock {Ph.D.} diss., Dept.\ of Computer Science, Stanford Univ., Stanford, Calif.

\bibitem[{Clancey(1983)}]{c:83}
Clancey, W.~J. 1983.
\newblock {Communication, Simulation, and Intelligent Agents: Implications of Personal Intelligent Machines for Medical Education}.
\newblock In \emph{Proceedings of the Eighth International Joint Conference on Artificial Intelligence {(IJCAI-83)}}, 556--560. Menlo Park, Calif: {IJCAI Organization}.

\bibitem[{Clancey(1984)}]{c:84}
Clancey, W.~J. 1984.
\newblock {Classification Problem Solving}.
\newblock In \emph{Proceedings of the Fourth National Conference on Artificial Intelligence}, 45--54. Menlo Park, Calif.: AAAI Press.

\bibitem[{Clancey(2021)}]{c:21}
Clancey, W.~J. 2021.
\newblock {The Engineering of Qualitative Models}.
\newblock Forthcoming.

\bibitem[{Engelmore and Morgan(1986)}]{em:86}
Engelmore, R.; and Morgan, A., eds. 1986.
\newblock \emph{Blackboard Systems}.
\newblock Reading, Mass.: Addison-Wesley.

\bibitem[{Filiot et~al.(2023)Filiot, Ghermi, Olivier, Jacob, Fidon, Mac~Kain, Saillard, and Schiratti}]{filiot2023scaling}
Filiot, A.; Ghermi, R.; Olivier, A.; Jacob, P.; Fidon, L.; Mac~Kain, A.; Saillard, C.; and Schiratti, J.-B. 2023.
\newblock Scaling self-supervised learning for histopathology with masked image modeling.
\newblock \emph{medRxiv}, 2023--07.

\bibitem[{Fourkioti, De~Vries, and Bakal(2023)}]{fourkioti2023camil}
Fourkioti, O.; De~Vries, M.; and Bakal, C. 2023.
\newblock CAMIL: Context-Aware Multiple Instance Learning for Cancer Detection and Subtyping in Whole Slide Images.
\newblock \emph{arXiv preprint arXiv:2305.05314}.

\bibitem[{Hasling, Clancey, and Rennels(1984)}]{hcr:83}
Hasling, D.~W.; Clancey, W.~J.; and Rennels, G. 1984.
\newblock Strategic explanations for a diagnostic consultation system.
\newblock \emph{International Journal of Man-Machine Studies}, 20(1): 3--19.

\bibitem[{Hasling et~al.(1983)Hasling, Clancey, Rennels, and Test}]{hcrt:83}
Hasling, D.~W.; Clancey, W.~J.; Rennels, G.~R.; and Test, T. 1983.
\newblock {Strategic Explanations in Consultation---Duplicate}.
\newblock \emph{The International Journal of Man-Machine Studies}, 20(1): 3--19.

\bibitem[{He et~al.(2020)He, Fan, Wu, Xie, and Girshick}]{he2020momentum}
He, K.; Fan, H.; Wu, Y.; Xie, S.; and Girshick, R. 2020.
\newblock Momentum contrast for unsupervised visual representation learning.
\newblock In \emph{Proceedings of the IEEE/CVF conference on computer vision and pattern recognition}, 9729--9738.

\bibitem[{Hou et~al.(2016)Hou, Samaras, Kurc, Gao, Davis, and Saltz}]{hou2016patch}
Hou, L.; Samaras, D.; Kurc, T.~M.; Gao, Y.; Davis, J.~E.; and Saltz, J.~H. 2016.
\newblock Patch-based convolutional neural network for whole slide tissue image classification.
\newblock In \emph{Proceedings of the IEEE conference on computer vision and pattern recognition}, 2424--2433.

\bibitem[{Huang et~al.(2023)Huang, Yao, Li, Mao, Huang, Hu, Hu, Wang, Guo, Tang et~al.}]{huang2023deep}
Huang, Y.; Yao, Z.; Li, L.; Mao, R.; Huang, W.; Hu, Z.; Hu, Y.; Wang, Y.; Guo, R.; Tang, X.; et~al. 2023.
\newblock Deep learning radiopathomics based on preoperative US images and biopsy whole slide images can distinguish between luminal and non-luminal tumors in early-stage breast cancers.
\newblock \emph{EBioMedicine}, 94.

\bibitem[{Ilse, Tomczak, and Welling(2018)}]{ilse2018attention}
Ilse, M.; Tomczak, J.; and Welling, M. 2018.
\newblock Attention-based deep multiple instance learning.
\newblock In \emph{International conference on machine learning}, 2127--2136. PMLR.

\bibitem[{Juyal et~al.(2024)Juyal, Shingi, Javed, Padigela, Shah, Sampat, Khosla, Abel, and Taylor-Weiner}]{juyal2024sc}
Juyal, D.; Shingi, S.; Javed, S.~A.; Padigela, H.; Shah, C.; Sampat, A.; Khosla, A.; Abel, J.; and Taylor-Weiner, A. 2024.
\newblock SC-MIL: Supervised Contrastive Multiple Instance Learning for Imbalanced Classification in Pathology.
\newblock In \emph{Proceedings of the IEEE/CVF Winter Conference on Applications of Computer Vision}, 7946--7955.

\bibitem[{Kanavati et~al.(2020)Kanavati, Toyokawa, Momosaki, Rambeau, Kozuma, Shoji, Yamazaki, Takeo, Iizuka, and Tsuneki}]{kanavati2020weakly}
Kanavati, F.; Toyokawa, G.; Momosaki, S.; Rambeau, M.; Kozuma, Y.; Shoji, F.; Yamazaki, K.; Takeo, S.; Iizuka, O.; and Tsuneki, M. 2020.
\newblock Weakly-supervised learning for lung carcinoma classification using deep learning.
\newblock \emph{Scientific reports}, 10(1): 9297.

\bibitem[{Lazard et~al.(2023)Lazard, Lerousseau, Decenci{\`e}re, and Walter}]{lazard2023giga}
Lazard, T.; Lerousseau, M.; Decenci{\`e}re, E.; and Walter, T. 2023.
\newblock Giga-ssl: Self-supervised learning for gigapixel images.
\newblock In \emph{Proceedings of the IEEE/CVF Conference on Computer Vision and Pattern Recognition}, 4305--4314.

\bibitem[{Lee et~al.(2022)Lee, Park, Oh, Shin, Sun, Jung, Lee, Kim, Chung, Moon et~al.}]{lee2022derivation}
Lee, Y.; Park, J.~H.; Oh, S.; Shin, K.; Sun, J.; Jung, M.; Lee, C.; Kim, H.; Chung, J.-H.; Moon, K.~C.; et~al. 2022.
\newblock Derivation of prognostic contextual histopathological features from whole-slide images of tumours via graph deep learning.
\newblock \emph{Nature Biomedical Engineering}, 1--15.

\bibitem[{Li, Li, and Eliceiri(2021)}]{li2021dual}
Li, B.; Li, Y.; and Eliceiri, K.~W. 2021.
\newblock Dual-stream multiple instance learning network for whole slide image classification with self-supervised contrastive learning.
\newblock In \emph{Proceedings of the IEEE/CVF conference on computer vision and pattern recognition}, 14318--14328.

\bibitem[{Li et~al.(2023)Li, Zhu, Zhang, Sun, Shui, Kuang, Zheng, and Yang}]{li2023task}
Li, H.; Zhu, C.; Zhang, Y.; Sun, Y.; Shui, Z.; Kuang, W.; Zheng, S.; and Yang, L. 2023.
\newblock Task-specific fine-tuning via variational information bottleneck for weakly-supervised pathology whole slide image classification.
\newblock In \emph{Proceedings of the IEEE/CVF Conference on Computer Vision and Pattern Recognition}, 7454--7463.

\bibitem[{Li et~al.(2019)Li, Wu, Wiliem, Zhao, Zhang, and Lovell}]{li2019deep}
Li, M.; Wu, L.; Wiliem, A.; Zhao, K.; Zhang, T.; and Lovell, B. 2019.
\newblock Deep instance-level hard negative mining model for histopathology images.
\newblock In \emph{Medical Image Computing and Computer Assisted Intervention--MICCAI 2019: 22nd International Conference, Shenzhen, China, October 13--17, 2019, Proceedings, Part I 22}, 514--522. Springer.

\bibitem[{Li et~al.(2018)Li, Yao, Zhu, Li, and Huang}]{li2018graph}
Li, R.; Yao, J.; Zhu, X.; Li, Y.; and Huang, J. 2018.
\newblock Graph CNN for survival analysis on whole slide pathological images.
\newblock In \emph{International Conference on Medical Image Computing and Computer-Assisted Intervention}, 174--182. Springer.

\bibitem[{Lin et~al.(2023)Lin, Yu, Xu, Hu, Xu, and Chen}]{lin2023sgcl}
Lin, T.; Yu, Z.; Xu, Z.; Hu, H.; Xu, Y.; and Chen, C.-W. 2023.
\newblock SGCL: Spatial guided contrastive learning on whole-slide pathological images.
\newblock \emph{Medical Image Analysis}, 89: 102845.

\bibitem[{Liu et~al.(2023)Liu, Fu, Ye, Yang, and Ji}]{liu2023dsca}
Liu, P.; Fu, B.; Ye, F.; Yang, R.; and Ji, L. 2023.
\newblock DSCA: A dual-stream network with cross-attention on whole-slide image pyramids for cancer prognosis.
\newblock \emph{Expert Systems with Applications}, 227: 120280.

\bibitem[{Lu et~al.(2023)Lu, Chen, Zhang, Williamson, Chen, Ding, Le, Chuang, and Mahmood}]{lu2023visual}
Lu, M.~Y.; Chen, B.; Zhang, A.; Williamson, D.~F.; Chen, R.~J.; Ding, T.; Le, L.~P.; Chuang, Y.-S.; and Mahmood, F. 2023.
\newblock Visual language pretrained multiple instance zero-shot transfer for histopathology images.
\newblock In \emph{Proceedings of the IEEE/CVF conference on computer vision and pattern recognition}, 19764--19775.

\bibitem[{Lu et~al.(2021)Lu, Williamson, Chen, Chen, Barbieri, and Mahmood}]{lu2021data}
Lu, M.~Y.; Williamson, D.~F.; Chen, T.~Y.; Chen, R.~J.; Barbieri, M.; and Mahmood, F. 2021.
\newblock Data-efficient and weakly supervised computational pathology on whole-slide images.
\newblock \emph{Nature biomedical engineering}, 5(6): 555--570.

\bibitem[{{NASA}(2015)}]{c:23}
{NASA}. 2015.
\newblock Pluto: The 'Other' Red Planet.
\newblock \url{https://www.nasa.gov/nh/pluto-the-other-red-planet}.
\newblock Accessed: 2018-12-06.

\bibitem[{Rice(1986)}]{r:86}
Rice, J. 1986.
\newblock {Poligon: A System for Parallel Problem Solving}.
\newblock Technical Report KSL-86-19, Dept.\ of Computer Science, Stanford Univ.

\bibitem[{Robinson(1980{\natexlab{a}})}]{r:80}
Robinson, A.~L. 1980{\natexlab{a}}.
\newblock New Ways to Make Microcircuits Smaller.
\newblock \emph{Science}, 208(4447): 1019--1022.

\bibitem[{Robinson(1980{\natexlab{b}})}]{r:80x}
Robinson, A.~L. 1980{\natexlab{b}}.
\newblock {New Ways to Make Microcircuits Smaller---Duplicate Entry}.
\newblock \emph{Science}, 208: 1019--1026.

\bibitem[{Shao et~al.(2021)Shao, Bian, Chen, Wang, Zhang, Ji et~al.}]{shao2021transmil}
Shao, Z.; Bian, H.; Chen, Y.; Wang, Y.; Zhang, J.; Ji, X.; et~al. 2021.
\newblock Transmil: Transformer based correlated multiple instance learning for whole slide image classification.
\newblock \emph{Advances in neural information processing systems}, 34: 2136--2147.

\bibitem[{Tang et~al.(2023)Tang, Huang, Zhang, Zhou, Zhang, and Liu}]{tang2023multiple}
Tang, W.; Huang, S.; Zhang, X.; Zhou, F.; Zhang, Y.; and Liu, B. 2023.
\newblock Multiple instance learning framework with masked hard instance mining for whole slide image classification.
\newblock In \emph{Proceedings of the IEEE/CVF International Conference on Computer Vision}, 4078--4087.

\bibitem[{Thandiackal et~al.(2022)Thandiackal, Chen, Pati, Jaume, Williamson, Gabrani, and Goksel}]{thandiackal2022differentiable}
Thandiackal, K.; Chen, B.; Pati, P.; Jaume, G.; Williamson, D.~F.; Gabrani, M.; and Goksel, O. 2022.
\newblock Differentiable zooming for multiple instance learning on whole-slide images.
\newblock In \emph{European Conference on Computer Vision}, 699--715. Springer.

\bibitem[{Vaswani et~al.(2017)Vaswani, Shazeer, Parmar, Uszkoreit, Jones, Gomez, Kaiser, and Polosukhin}]{c:22}
Vaswani, A.; Shazeer, N.; Parmar, N.; Uszkoreit, J.; Jones, L.; Gomez, A.~N.; Kaiser, L.; and Polosukhin, I. 2017.
\newblock Attention Is All You Need.
\newblock arXiv:1706.03762.

\bibitem[{Wang et~al.(2022)Wang, Xiang, Zhang, Yang, Yang, Wang, Zhang, Yang, Huang, and Han}]{wang2022scl}
Wang, X.; Xiang, J.; Zhang, J.; Yang, S.; Yang, Z.; Wang, M.-H.; Zhang, J.; Yang, W.; Huang, J.; and Han, X. 2022.
\newblock Scl-wc: Cross-slide contrastive learning for weakly-supervised whole-slide image classification.
\newblock \emph{Advances in neural information processing systems}, 35: 18009--18021.

\bibitem[{Xiang and Zhang(2023)}]{xiang2023exploring}
Xiang, J.; and Zhang, J. 2023.
\newblock Exploring low-rank property in multiple instance learning for whole slide image classification.
\newblock In \emph{The Eleventh International Conference on Learning Representations}.

\bibitem[{Yang et~al.(2023)Yang, Tu, Lei, and Long}]{yang2023hamil}
Yang, Y.; Tu, Y.; Lei, H.; and Long, W. 2023.
\newblock HAMIL: Hierarchical aggregation-based multi-instance learning for microscopy image classification.
\newblock \emph{Pattern Recognition}, 136: 109245.

\bibitem[{Yu et~al.(2020)Yu, Deng, Liu, Zhou, Jia, Xiao, Zhou, Li, Guo, Wang et~al.}]{yu2020lymph}
Yu, J.; Deng, Y.; Liu, T.; Zhou, J.; Jia, X.; Xiao, T.; Zhou, S.; Li, J.; Guo, Y.; Wang, Y.; et~al. 2020.
\newblock Lymph node metastasis prediction of papillary thyroid carcinoma based on transfer learning radiomics.
\newblock \emph{Nature communications}, 11(1): 4807.

\bibitem[{Yu et~al.(2023)Yu, Ma, Fu, Chen, Lai, Zhuo, and Xu}]{yu2023local}
Yu, J.; Ma, T.; Fu, Y.; Chen, H.; Lai, M.; Zhuo, C.; and Xu, Y. 2023.
\newblock Local-to-global spatial learning for whole-slide image representation and classification.
\newblock \emph{Computerized Medical Imaging and Graphics}, 107: 102230.

\bibitem[{Zhang et~al.(2022)Zhang, Meng, Zhao, Qiao, Yang, Coupland, and Zheng}]{zhang2022dtfd}
Zhang, H.; Meng, Y.; Zhao, Y.; Qiao, Y.; Yang, X.; Coupland, S.~E.; and Zheng, Y. 2022.
\newblock Dtfd-mil: Double-tier feature distillation multiple instance learning for histopathology whole slide image classification.
\newblock In \emph{Proceedings of the IEEE/CVF conference on computer vision and pattern recognition}, 18802--18812.

\bibitem[{Zhang et~al.(2023)Zhang, Kapse, Ma, Prasanna, Saltz, Vakalopoulou, and Samaras}]{zhang2023prompt}
Zhang, J.; Kapse, S.; Ma, K.; Prasanna, P.; Saltz, J.; Vakalopoulou, M.; and Samaras, D. 2023.
\newblock Prompt-mil: Boosting multi-instance learning schemes via task-specific prompt tuning.
\newblock In \emph{International Conference on Medical Image Computing and Computer-Assisted Intervention}, 624--634. Springer.

\bibitem[{Zhao et~al.(2020)Zhao, Yang, Fang, Liu, Zhou, Zhang, Sun, Yang, Menze, Fan et~al.}]{zhao2020predicting}
Zhao, Y.; Yang, F.; Fang, Y.; Liu, H.; Zhou, N.; Zhang, J.; Sun, J.; Yang, S.; Menze, B.; Fan, X.; et~al. 2020.
\newblock Predicting lymph node metastasis using histopathological images based on multiple instance learning with deep graph convolution.
\newblock In \emph{Proceedings of the IEEE/CVF conference on computer vision and pattern recognition}, 4837--4846.

\bibitem[{Zheng et~al.(2022)Zheng, Gindra, Green, Burks, Betke, Beane, and Kolachalama}]{zheng2022graph}
Zheng, Y.; Gindra, R.~H.; Green, E.~J.; Burks, E.~J.; Betke, M.; Beane, J.~E.; and Kolachalama, V.~B. 2022.
\newblock A graph-transformer for whole slide image classification.
\newblock \emph{IEEE transactions on medical imaging}, 41(11): 3003--3015.

\bibitem[{Zhu et~al.(2023)Zhu, Yu, Wu, Yu, Zhang, and Wang}]{zhu2023murcl}
Zhu, Z.; Yu, L.; Wu, W.; Yu, R.; Zhang, D.; and Wang, L. 2023.
\newblock MuRCL: Multi-Instance Reinforcement Contrastive Learning for Whole Slide Image Classification.
\newblock \emph{IEEE Transactions on Medical Imaging}, 42(5): 1337--1348.

\end{thebibliography}

\end{document}